\documentclass[10pt,twocolumn,letterpaper]{article}

\usepackage[pagenumbers]{iccv} 

\usepackage{microtype}
\usepackage{arydshln}
\newcommand{\Fref}[1]{Fig.~\ref{#1}}
\newcommand{\Tref}[1]{Table~\ref{#1}}

\definecolor{iccvblue}{rgb}{0.21,0.49,0.74}
\usepackage[pagebackref,breaklinks,colorlinks,allcolors=iccvblue]{hyperref}

\def\paperID{6184} 
\def\confName{ICCV}
\def\confYear{2025}

\title{VSC: Visual Search Compositional Text-to-Image Diffusion Model}

\begin{document}
\author{
    Do Huu Dat\textsuperscript{1*} \quad
    Nam Hyeonu\textsuperscript{2} \quad
    Po-Yuan Mao\textsuperscript{3} \quad
    Tae-Hyun Oh\textsuperscript{2,4}
}

\affil{\textsuperscript{1}VinUniversity \quad
       \textsuperscript{2}POSTECH \quad
       \textsuperscript{3}Academia Sinica \quad
       \textsuperscript{4}KAIST}

\twocolumn[{%
\renewcommand\twocolumn[1][]{#1}%
\maketitle
\begin{center}
    \centering
    \captionsetup{type=figure}
    \includegraphics[width=0.9\textwidth]{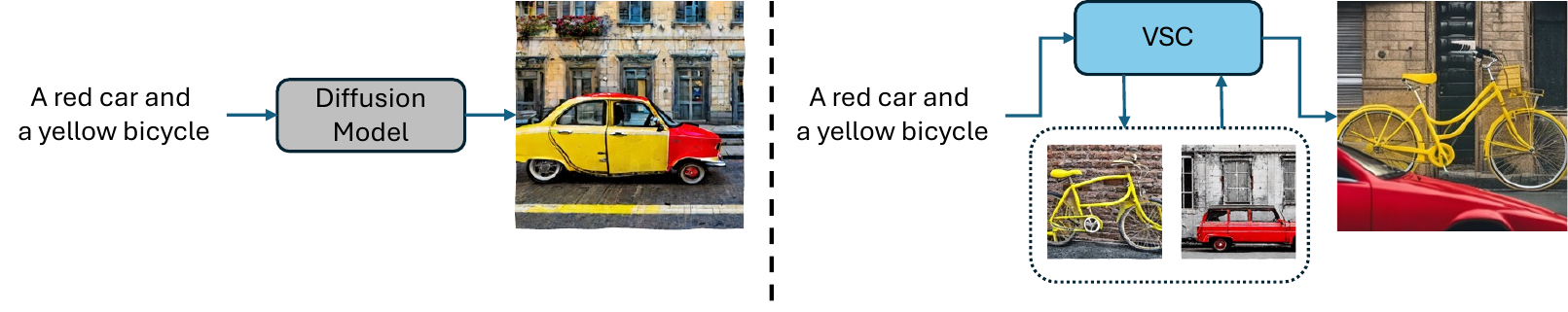}
    \captionof{figure}{\textbf{Left:} Original pipeline and generated image of Diffusion Model. \textbf{Right:} VSC individually ``search'' for visual information of each binding pair by generating images for each separate pair (``red car'' and ``yellow bicycle''). Finally, VSC uses them as references to generate the correct compositional image.}
    \label{problem}
\end{center}%
}]
\def\thefootnote{*}\footnotetext{Work is done as an intern at POSTECH}\def\thefootnote{\arabic{footnote}}
\begin{abstract}
Text-to-image diffusion models have shown impressive capabilities in generating realistic visuals from natural-language prompts, yet they often struggle with accurately binding attributes to corresponding objects, especially in prompts containing multiple attribute-object pairs. This challenge primarily arises from the limitations of commonly used text encoders, such as CLIP, which can fail to encode complex linguistic relationships and modifiers effectively. Existing approaches have attempted to mitigate these issues through attention map control during inference and the use of layout information or fine-tuning during training, yet they face performance drops with increased prompt complexity. In this work, we introduce a novel compositional generation method that leverages pairwise image embeddings to improve attribute-object binding. Our approach decomposes complex prompts into sub-prompts, generates corresponding images, and computes visual prototypes that fuse with text embeddings to enhance representation. By applying segmentation-based localization training, we address cross-attention misalignment, achieving improved accuracy in binding multiple attributes to objects. Our approaches outperform existing compositional text-to-image diffusion models on the benchmark T2I CompBench, achieving better image quality, evaluated by humans, and emerging robustness under scaling number of binding pairs in the prompt.
\end{abstract}    
\section{Introduction}
\label{sec:intro}
Compositionality is a fundamental ability that enables humans to naturally comprehend the world. It allows us to quickly recognize new objects through their components or construct complex sentences by combining words. This skill empowers us to make ``infinite use of finite mean'' \cite{chomsky2014aspects}, endlessly reusing and recombining familiar concepts in diverse ways. In terms of developing compositional AI systems, diffusion models rise as the most feasible component due to their inherent enormous mode coverage and flexible controllability. The diffusion model's possession of compositional generalization has been proved with synthetic data and the dataset CelebA, where the images are easily decomposed into gender and race \cite{okawa2024compositional}. Additionally, diffusion models for generating images from text prompts are capable of creating remarkably realistic visuals, allowing users to shape the content of generated images by crafting detailed and complex natural-language prompts. 

Despite these capabilities, these models often fail to fully utilize compositionality and produce inaccurate images when provided with more intricate descriptions. A prevalent issue is improper binding, as illustrated in the left side of \Fref{problem}, where descriptive words fail to correctly influence the visual characteristics of specific nouns they are intended to modify. For instance, the prompt specifying “A photo of a red car and a yellow bicycle” may produce an image with mixed or misplaced attributes. One likely cause for these inconsistencies is the reliance on text encoders like CLIP within diffusion models. 
Although CLIP demonstrates strong performance in aligning images with textual descriptions, it faces significant challenges in handling complex linguistic structures. Specifically, it struggles with encoding hierarchical relationships and modifiers, which are crucial for understanding nuanced textual inputs. Previous studies \cite{lewis2022does, tong2024eyes, toker2024diffusion} have highlighted that CLIP tends to simplify textual inputs by representing them as a ``bag-of-concepts,'' disregarding the intricate dependencies and relationships between words. These inherent weaknesses pose limitations for diffusion models that rely on CLIP for text encoding, as they hinder the model's capacity to fully capture and respect the precise linguistic relationships specified in prompts.

Serial search theory \cite{treisman1980feature} in cognitive psychology suggests that when humans encounter objects with multiple features (e.g., color and shape), our brain processes them individually to bind these features into a cohesive perception. The theory has been successfully applied to improve the capability of large vision-language models \cite{campbell2024understanding}. Noticeably, CLIP likewise strongly materializes compositionality in the single binding setting \cite{lewis2022does} and this property is well-preserved in the latent diffusion model (refer to Section \ref{single-compo}). Providing spatial information such as bounding box, segmentation maps, blobs, \cite{dahary2024yourself, nie2024compositional, zheng2023layoutdiffusion, feng2024layoutgpt} can drastically improve the compositional generation by controlling the attention to be bounded in a specific regions while maintaining a reasonable layout. Those methods can be interpreted as disentangling the desired image into object-level generation. This leads to our motivation to generate images from every single attribute-object pair as references for improving the compositional generalization of the text-to-image diffusion model.

On the other hand, the diffusion model can merge and edit images, namely subject-driven image generation, by quickly fine-tuning the model \cite{ruiz2023dreambooth} that encodes a unique identifier that refers to the subject. Notably, Xiao et al. \cite{xiao2024fastcomposer} propose a tuning-free method via training the diffusion model alongside an image encoder and an MLP. Besides, training an MLP has also been shown to be effective in improving the compositionality of the text-to-image model. Therefore, taking the psychological theory as inspiration, with the above previous works in diffusion models, our key idea, illustrated on the right side of Fig.~\ref{problem}, is to develop a framework that generates images for each binding pair individually and ultimately fuses these images into a complete image from a given prompt containing several attribute-object binding pairs. The fusing process requires the use of an external MLP layer, so we create a synthetic dataset and fine-tune the MLP. Moreover, we extract the segmentation maps of the object and deploy cross-attention localization training as in \cite{xiao2024fastcomposer} to fix the erroneous cross-attention maps, aiming to address the extreme decrease while increasing the number of attribute-object pairs.


Our main contributions are summarized as follows:
\begin{itemize}
    \item We present VSC, a novel compositional text-to-image diffusion generation method based on a tuning-free, subject-driven image generation framework. Remarkably, VSC requires training only an MLP and the final layers of the image encoder.
    \item We leverage the remarkable ability of the pre-trained text-to-image diffusion model in handling single attribute binding prompts to create synthetic images as reference images throughout training and inference.
    \item We achieve state-of-the-art performance on the attribute binding benchmark compared to other compositional text-to-image diffusion models, and provide extensive analysis as well as ablation studies.
\end{itemize}
\section{Related Work}

\textbf{Compositional Text-to-Image Diffusion Model}: In recent work, Huang et al. \cite{huang2023t2i} introduce a benchmark to test compositionality in text-to-image models, highlighting the vulnerability of open-source models to simple compositional prompts. They also propose a fine-tuning approach to enhance compositionality in these models. Several existing studies suggest different methods for enhancing compositionality in text-to-image models in both inference time and training time. In terms of inference time, one popular way is controlling the attention maps of the cross-attention layers \cite{rassin2024linguistic, wang2024compositional, meral2024conform, agarwal2023star}, which would ensure the maps of binding tokens are overlapping. However, because the attention scores are illustrated to be erroneous \cite{zarei2024understanding}, then scaling the number of pairs leads to a drastic decrease in performance in those works.  Furthermore, combining the guidance in a structured approach \cite{liu2022compositional, feng2022training} also mitigates the binding problems. 

During training time, other works leverage the layout information (bounding box, segmentation)\cite{feng2024layoutgpt, nie2024compositional, zheng2023layoutdiffusion} to correctly represent the spatial information of the image and fine-tune the diffusion model to concentrate on the binding problem. Additionally, \cite{zarei2024understanding, huang2023t2i} demonstrate the efficiency of fine-tuning the UNet or an MLP layer, which transforms the text embeddings. Our work aims to resolve the discrepancy between cross-attention maps and propose a new approach to tackling compositional generalization without the need for human-annotated layouts or direct latent space manipulation.
\begin{figure*}[htb]
    \centering
    \includegraphics[width=\textwidth]{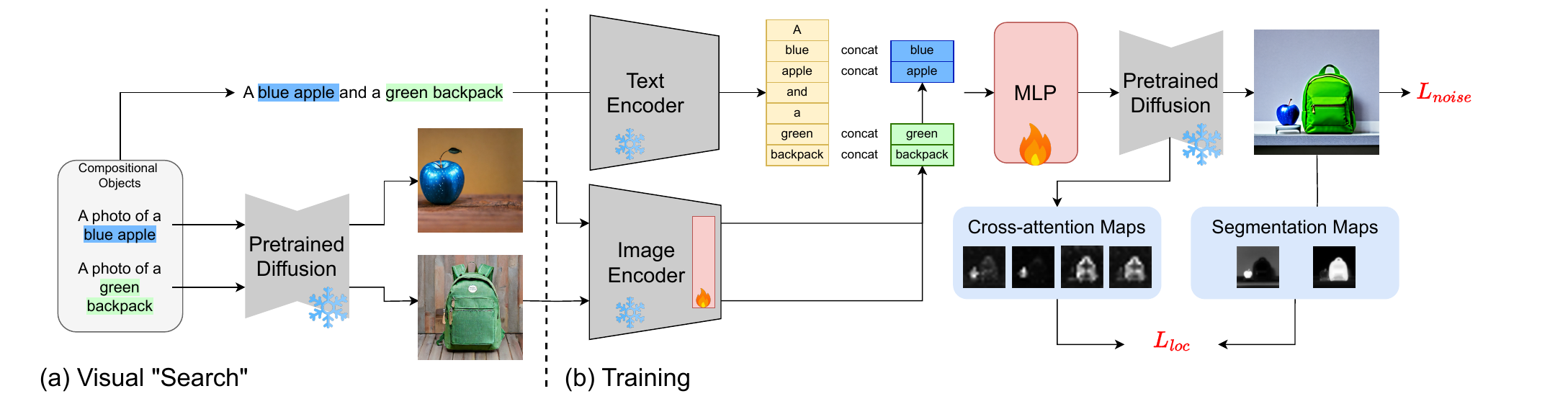}
    \caption{\textbf{The training pipeline of VSC.} Given the input text prompt ``A blue apple and a green backpack'', VSC first generates images of each binding pair (``blue apple'' and ``green backpack''), then uses an image encoder to extract visual prototype features, which is later used to augment the text embeddings via a trainable MLP module. Finally, a frozen pre-trained diffusion model generates the image with the augmented text embeddings. Additionally, we compute the $L_{loc}$ loss between cross-attention maps and segmentation maps to enhance the precision of cross-attention maps.}
    \label{method}
    \vspace{-1em}
\end{figure*}

\noindent \textbf{Subject-driven Image Generation}: Text-to-image personalization is a growing application where models are fine-tuned to generate unique renditions of subjects based on a few reference images \cite{gal2022image, kumari2023multi, ruiz2023dreambooth}. Notable works like DreamBooth \cite{ruiz2023dreambooth} and Custom Diffusion \cite{kumari2023multi} excel in this area, generating highly personalized outputs by fine-tuning a pre-trained text-to-image diffusion model with a few reference images and a textual prompt in the form of ``a [v] [class name]'' or ``photo of a [v] [class name].'' In this format, [v] serves as the subject's unique identifier—usually a rare token with minimal inherent meaning—while [class name] represents a broad category for the subject. 

Recent approaches to subject-driven image generation focus on synthesizing new images of specific subjects using only a few reference photos but often require computationally expensive tuning, such as DreamBooth \cite{ruiz2023dreambooth}, Textual Inversion \cite{mokady2023null}, and Custom Diffusion \cite{kumari2023multi}. Some methods, like Tuning Encoder \cite{gal2023encoder}, aim to reduce fine-tuning steps by creating initial latent codes with pre-trained encoders but still face scalability challenges. Tuning-free alternatives like X\&Fuse \cite{kirstain2023x}, ELITE \cite{wei2023elite}, and InstantBooth \cite{shi2024instantbooth} use embedding techniques to map reference images into text conditioning, although these methods primarily support single-subject generation. For multi-subject scenarios, Custom Diffusion \cite{kumari2023multi} employs joint tuning, while others, like SpaText \cite{avrahami2023spatext} and Collage Diffusion \cite{sarukkai2024collage}, rely on segmentation masks to define object layouts. However, these techniques struggle with subjects of similar types or require user-specified layouts, which limit flexibility. FastComposer \cite{xiao2024fastcomposer} overcomes these limitations by implementing cross-attention localization for better subject separation. To the best of our knowledge, we are the first to utilize subject-driven image generation in solving attribute-object binding problems, fully operating on synthetic images as reference images.
\section{Our Approach}

\subsection{Preliminaries}

We build upon the Stable Diffusion (SD) model, a state-of-the-art framework comprising three main components: a variational autoencoder (VAE), a U-Net, and a text encoder. In this setup, the VAE encoder $E$ compresses an image $x$ into a smaller latent space representation $z$, which is then perturbed with Gaussian noise $\varepsilon$ during the forward diffusion process. The U-Net, parameterized by $\theta$, is trained to denoise the latent representation by predicting and removing the noise. Text prompts condition this denoising process via a cross-attention mechanism, where the text encoder $\psi$ maps prompts $P$ to embeddings $\psi(P)$. During training, the model minimizes a noise loss function defined as:
\begin{equation}
L_{\text{noise}} = \mathbb{E}_{z \sim \mathcal{E}(x), P, \varepsilon \sim \mathcal{N}(0, 1), t} 
\left[
\|\varepsilon - \varepsilon_\theta(z_t, t, \psi(P))\|_2^2
\right],
\end{equation}
where $z_t$ is the latent variable at time step $t$. At inference, a noise sample $z_T$ is drawn from $\mathcal{N}(0, 1)$ and gradually refined by the U-Net to recover $z_0$, after which the VAE decoder $D$ reconstructs the final image $ \hat{x} = D(z_0)$.

The cross-attention mechanism in the U-Net conditions the denoising process on the text prompt. This is achieved by transforming the latent code $z \in \mathbb{R}^{(h \times w) \times f}$ and text embeddings $c \in \mathbb{R}^{n \times d}$, where the text encoder $\psi$ converts the text $P$ into a set of $d$-dimensional embeddings. Both the latent code and the text embeddings are projected into Query, Key, and Value matrices: $Q = W^q z$, $K = W^k c$, and $V = W^v c$, where $W^q \in \mathbb{R}^{f \times d'}$ and $W^k, W^v \in \mathbb{R}^{d \times d'}$ are the learned projection matrices, and $d'$ is the shared dimension of Query, Key, and Value. The attention scores are computed as:

\begin{equation}
A = \text{Softmax}\left(\frac{QK^\top}{\sqrt{d'}}\right) \in [0, 1]^{(h \times w) \times n}
\end{equation}

allowing the model to compute a weighted sum over the Value matrix, resulting in the cross-attention output $z_{\text{attn}} = AV \in \mathbb{R}^{(h \times w) \times d'}$. This mechanism distributes textual information across the latent spatial dimensions, where $A[i, j, k]$ denotes the information flow from the $k$-th text token to the latent pixel at position $(i, j)$. This cross-attention structure provides a foundation for interpreting semantic information in the latent code.

\subsection{Visual Embedding Fusion}
Given a prompt $P$ including multiple binding pairs $[[a_1, o_1], [a_2, o_2], \ldots, [a_n, o_n]]$ having the corresponding index $\mathcal{I} = [i_{a1}, i_{o1}, i_{a2}, i_{o2}, \ldots, i_{an}, i_{on}]$, a pretrained image encoder $\phi: \mathbb{R}^{C \times H \times W} \xrightarrow{} \mathbb{R}^{D}$ and a lightweight MLP$: \mathbb{R}^{2*D} \xrightarrow{} \mathbb{R}^{D}$. We first extract the pairs and generate $m$ images from each pair $[a_n, o_n]$ using SD, so we have the corresponding set of reference images $[[r_1^1, \ldots r_1^m], \ldots, [r_n^1, \ldots, r_n^m]]$. Then, the set of images is fed into the image encoder $\phi$ to produce the image embeddings and we take the mean of each set to get the prototype representation of each binding pair according to the following equations:
\begin{align}
    \begin{split}
        v_j^k = \phi(r_j^k)
    \end{split}
    \\
    \begin{split}
        \centering
        e_j = \frac{1}{m} \sum_{k=0}^{m} v_j^k \quad \text{for} \quad j \in \mathcal{I} 
    \end{split}
\end{align}

\noindent resulting in the visual prototype features $[e_1, e_2, \ldots, e_n]$. Finally, we augment the text embeddings $c$ with those visual prototype features. For each token in c: 
\begin{align}
    c'_i = \begin{cases} 
    c_i, & i \notin I \\ 
    \operatorname{MLP}([c_i, e_j]), & i = i_j \in I
    \end{cases}
\end{align}
The cross-attention maps in the original Stable Diffusion are erroneous, leading to incorrect binding (refer to Section.~\ref{abl-attn}). Therefore, we adopt the localizing cross-attention loss from FastComposer \cite{xiao2024fastcomposer} with minor modifications enhancing the alignment between the cross-attention maps of attribute and object. Let \( \mathcal{M} = \{ M_1, M_2, \dots, M_n \} \) be the segmentation masks for the reference pairs existing in the denoising target. Define \( A_{i_{a}} = A[:, :, i_{a}], A_{i_{o}} = A[:, :, i_{o}] \in [0, 1]^{h \times w} \) as the cross-attention maps of the \( i \)-th pairs, comprising one map for the attribute and another map for the object. We then enforce the both 2 maps $A_{i_{a}}, A_{i_{o}}$ to approximate the segmentation mask \( M_j \) for the \( j \)-th subject token, i.e., \( A_{i_{a}} \approx A_{i_{o}} \approx M_j \). We use a balanced \( L_1 \) loss that minimizes the difference between the cross-attention map and the segmentation mask:
\begin{align}
    L_{\text{loc}} &= \frac{1}{n} \sum_{j=1}^n \bigg[ 
    \big(\text{mean}(A_{i_a} \odot \bar{M}_j) - \text{mean}(A_{i_a} \odot M_j)\big) \nonumber \\
    &\quad + \big(\text{mean}(A_{i_o} \odot \bar{M}_j) - \text{mean}(A_{i_o} \odot M_j)\big) \bigg], 
\end{align}
where $\text{mean}(A) = \frac{1}{h \cdot w} \sum_{i=1}^h \sum_{k=1}^w A_{ik}$. The overall training objective is then defined as:
\[
L = L_{\text{noise}} + \lambda L_{\text{loc}}.
\]


\subsection{Dataset Creation}
We make use of the T2I-CompBench dataset \cite{huang2023t2i}, concentrating on three attribute categories—color, texture, and shape—each containing 700 prompts for training. To generate high-quality images, we use three generative models: Stable Diffusion 3.5 \cite{rombach2022high}, Flux, and SynGen \cite{rassin2024linguistic} to generate 300 images for each prompt. Then we use Mask2Former \cite{cheng2021mask2former} for instance segmentation and OpenCLIP \cite{ilharco_gabriel_2021_5143773} to compute the alignment score of the segmentation and the given attribute-object exists in the prompt. Specifically, we run a greedy algorithm to assign the mask with the text embedding that has the highest alignment score. Additionally, our evaluation emphasizes the disentangled BLIP-Visual Question Answering (VQA) score introduced by Huang et al. \cite{huang2023t2i} as a metric for image quality. The VQA score measures how precisely an image captures the compositional elements specified in a prompt and aligns more closely with human judgment compared to metrics like CLIP-Score \cite{radford2021learning}. Finally, we select 45 images with the best scores for each prompt, resulting in the final dataset consisting of 90,000 images and their precise segmentation.
\section{Experiments}

\subsection{Implementation Details}

We use Stable Diffusion 1.4, 2.1, and 3.5~\citep{rombach2022high, esser2024scaling}.
The MLP module is composed of 2 linear layers with a LayerNorm.
For the image encoder, we use OpenAI's $\texttt{clip-vit-large-patch14}$ for Stable Diffusion 1.4 and LAION's $\texttt{CLIP-ViT-H-14}$ for Stable Diffusion 2.1 and Stable Diffusion 3.5. 
Note that we only optimize the last few layers in the image encoders.
We set the number of training steps to $60,000$ steps, the learning rate to $1e-5$, and the batch size to 8. 
We use 2 NVIDIA A5000 GPUs, and the training takes approximately 45 hours.

\subsection{Evaluation Results}
\label{eval_sec}

\textbf{How well can VSC compose the object with its corresponding attribute?} 
\citet{huang2023t2i} has proposed the BLIP-vqa score to evaluate the composition capability. We adopt their evaluation method on attributes of color, shape, and texture. Furthermore, we compute the harmonic mean (HM) to unify the scores into one as follows:
\begin{equation*}
    \text{HM} = \frac{3}{\frac{1}{C} + \frac{1}{T} + \frac{1}{S}},
\end{equation*}
where \( C \), \( T \), and \( S \) represent the individual performance scores for Color, Texture, and Shape, respectively.
We evaluate the methods on T2I CompBench dataset.

\begin{table}[!htb]
\centering
\begin{tabular}{lcccc}
\toprule
\textbf{Model} & \textbf{Color} & \textbf{Texture} & \textbf{Shape} & \textbf{HM}\\
\midrule
\multicolumn{5}{c}{Stable Diffusion 1.4} \\
\hdashline
Baseline      & 0.37 & 0.41 & 0.35 & 0.375\\
WiCLP & 0.53 & 0.56 & 0.45 & 0.509 \\
SynGen & 0.63 & 0.57 & 0.46 & 0.544 \\
\textbf{VSC} & \textbf{0.66} & \textbf{0.61} & \textbf{0.47} & \textbf{0.568}\\
\midrule
\multicolumn{5}{c}{Stable Diffusion 2.1} \\
\hdashline
Baseline      & 0.50 & 0.49 & 0.42 & 0.467\\
Composable & 0.40 & 0.36 & 0.32 & 0.357 \\
Structured  & 0.49 & 0.49 & 0.42 & 0.464 \\
Attn-Exct & 0.64 & 0.59 & 0.45 & 0.547 \\
GORS-unbiased & 0.64 & 0.60 & 0.45 & 0.550\\
WiCLP & 0.65 & 0.60 & 0.48 & 0.567 \\
CONFORM & 0.68 & 0.61 & 0.49 & 0.5824 \\
InitNO & 0.69 & 0.61 & 0.49 & 0.5849 \\
SynGen & 0.71 & 0.61 & 0.50 & 0.594 \\
\textbf{VSC} & \textbf{0.74} & \textbf{0.64} & \textbf{0.53} & \textbf{0.608}\\ 
\hline
\multicolumn{5}{c}{Stable Diffusion 3.5} \\
\hdashline
Baseline & 0.76 & 0.67 & 0.53 & 0.639\\
Attn-Exct & 0.79 & 0.69 & 0.53 & 0.652 \\
CONFORM & 0.80 & 0.69 & 0.53 & 0.654\\
InitNO & 0.79 & 0.70 & 0.54 & 0.660\\
SynGen & 0.82 & 0.74 & 0.59 & 0.703 \\
\textbf{VSC} & \textbf{0.85} & \textbf{0.79} & \textbf{0.63} & \textbf{0.727}\\ 
\bottomrule
\end{tabular}
\caption{\textbf{Quantitative results on T2I CompBench.} We compare the state-of-the-art methods with ours on 3 categories of the benchmark dataset (color, texture, and shape). Additionally, we provide the harmonic mean. With the same backbone, VSC is consistently better than others. The best is Stable Diffusion 3.5 with VSC. 
}
\vspace{-2em}
\label{table:comparison}
\end{table}

\Tref{table:comparison} lists the performance of baseline, ours, and previous work, such as Composable \cite{liu2022compositional} and Structured \cite{feng2022training}.
In Stable Diffusion 1.4, our method achieves the highest scores across all three attributes (Color: 0.66, Texture: 0.61, Shape: 0.47), outperforming the other methods. In Stable Diffusion 2.1, our method also achieves the best performance, with scores of 0.74 for Color, 0.64 for Texture, and 0.53 for Shape, indicating significant improvements over other models like GORS-unbiased \cite{huang2023t2i}, WiCLIP \cite{zarei2024understanding}, and SynGen \cite{rassin2024linguistic}. 
Stable Diffusion 3.5 with VSC archives the highest performance, resulting in 0.85, 0.79, and 0.63 for Color, Texture, and Shape, respectively.
The results 
demonstrates the effectiveness of our method in generating images with accurate color, texture, and shape fidelity across all three versions of Stable Diffusion. Our method can be applied to any diffusion model.

\noindent \textbf{Does VSC's performance drop with an increasing number of binding pairs in prompts?} 
Previous cross-attention control methods \cite{chefer2023attend, rassin2024linguistic, meral2024conform} often experience a performance drop with an increasing number of binding pairs in prompts. Rassin et al \cite{rassin2024linguistic} hypothesize that the loss in those cases becomes too large, making the image start to lose its visual quality. Given that the objective of those methods is to separate cross-attention maps, we believe that erroneous maps, as shown in \cite{zarei2024understanding}, are the main factor causing failure. Thus, with the integration of the localizing cross-attention, we expect the cross-attention maps of VSC to be more precise and less affected by the aforementioned phenomenon. 
\begin{table}[!htb]
    \centering
    \begin{tabular}{c c c c}
        \hline
        Method & 3 & 4 & 5 \\
        \hline
        \multicolumn{4}{c}{Stable Diffusion 2.1} \\
        \hdashline
         SynGen & 0.6779 & 0.4846 & 0.1189 \\
         VSC & \textbf{0.6881} & \textbf{0.5479} & \textbf{0.1901} \\
         \hline
         \multicolumn{4}{c}{Stable Diffusion 3.5} \\
         \hdashline
         SynGen & 0.8081 & 0.5664 & 0.1815 \\
         VSC & \textbf{0.8132} & \textbf{0.6183} & \textbf{0.2461} \\
         \hline
    \end{tabular}
    \caption{\textbf{Quantitative score with Different Numbers of Binding Pairs.} We randomly combine attributes and objects from T2I CompBench into single prompts to analyze compositions with varying complexity. Our method consistently performs well, even when handling a high number of bindings.
    }
    \label{scaling_obj}
    \vspace{-2em}
\end{table}
Thus, we evaluate VSC's performance under incrementing numbers of binding pairs. To set up the experiments, we provide all the individual attributes and objects, used in the T2I-CompBench validation set, and randomly combine them into the prompt with the template ``a [attribute] [object], ..., a [attribute] [object]'' with the desired number of pairs. Finally, we use chatGPT to correct the grammar of the prompt and avoid misleading of index in the fusing computation.

\begin{figure}[!htb]
    \centering
    \includegraphics[width=\columnwidth]{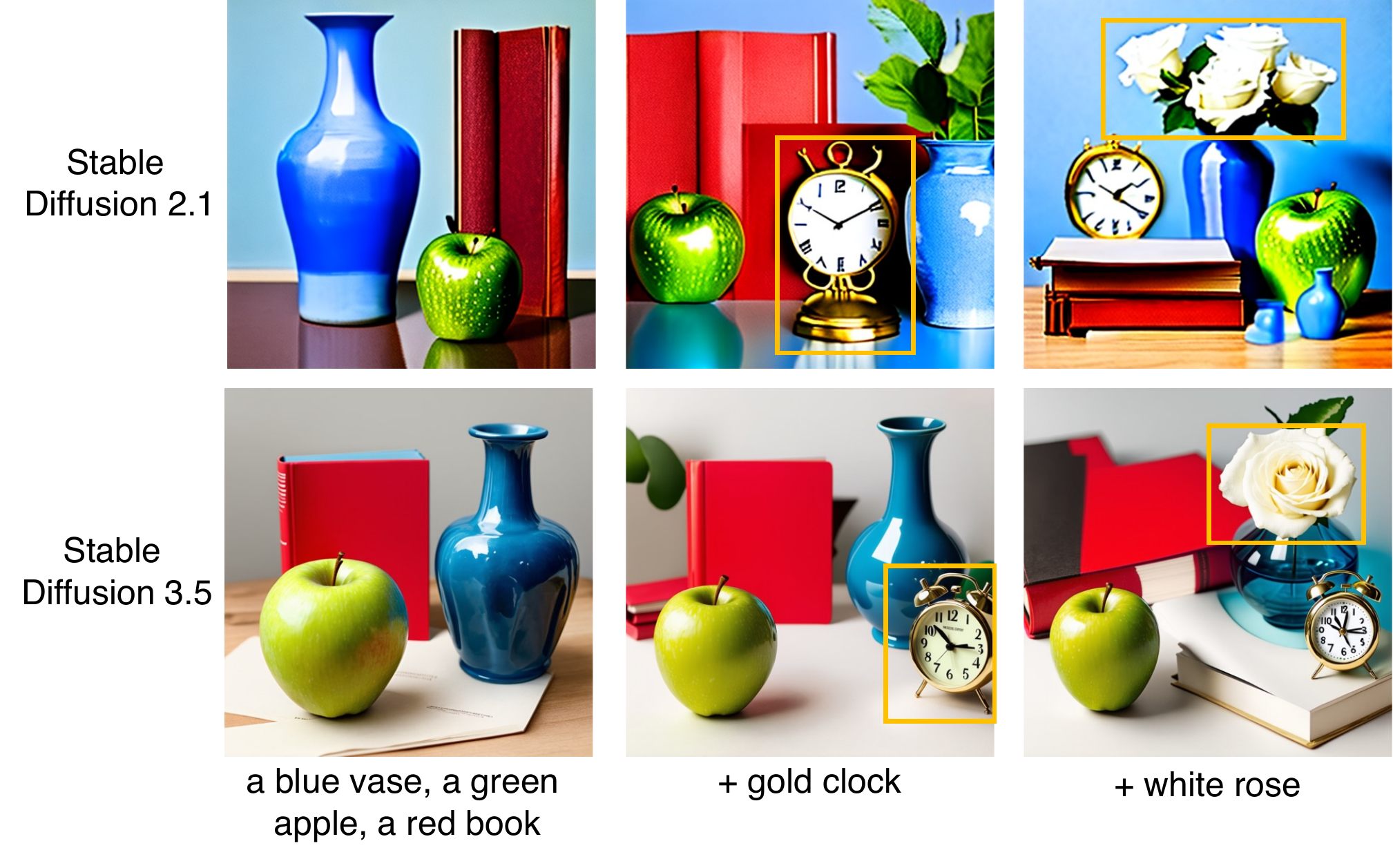}
    \caption{\textbf{Generation with the increasing number of binding pairs.} We show the generated images when the prompt includes more pairs. We observe that the generated image consistently reflects the additional composition.
    }
    \label{quali-obj-scale}
\end{figure}

\begin{figure*}[!htb]
    \centering
    \includegraphics[width=0.97\linewidth]{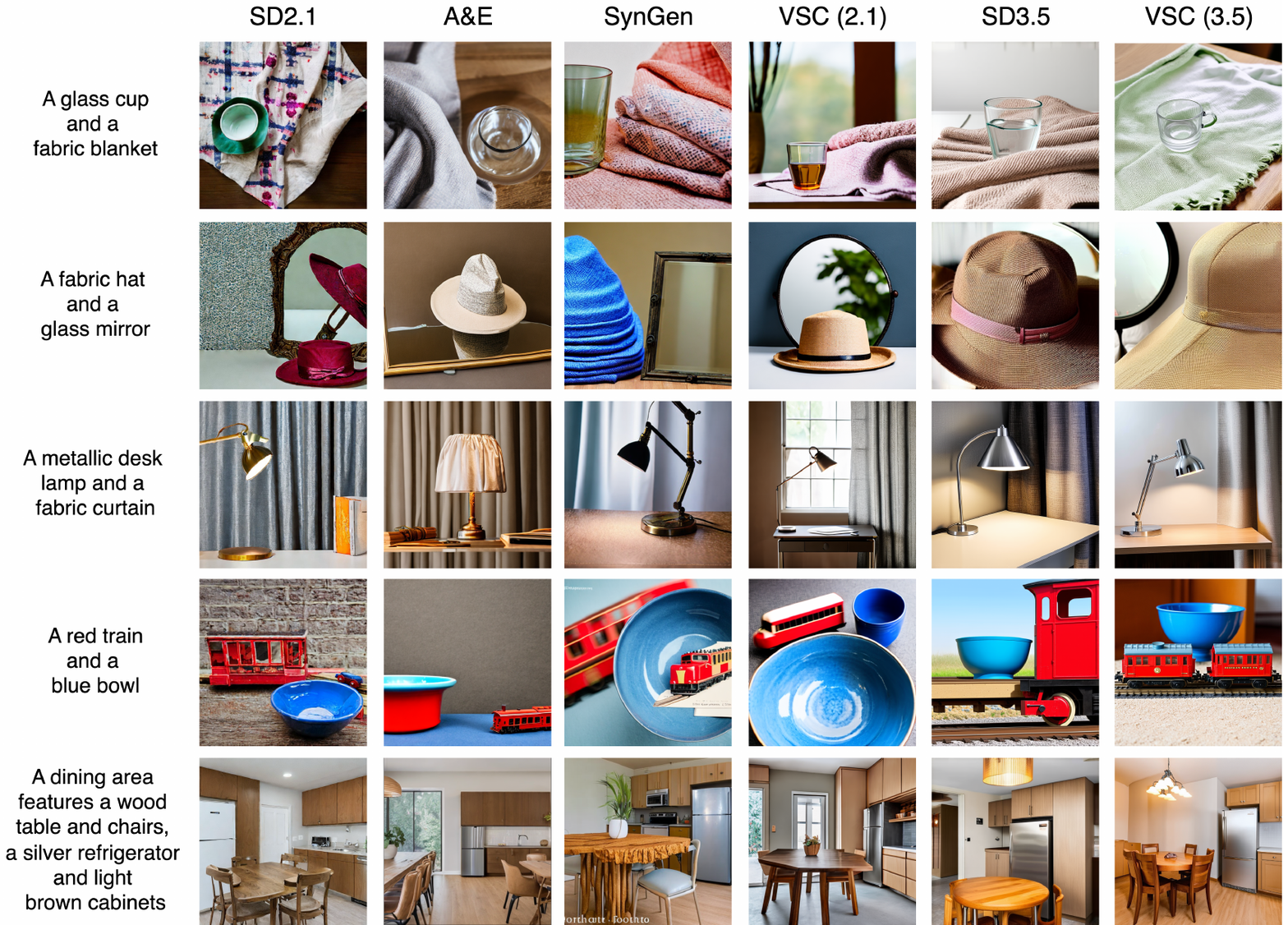}
    \caption{\textbf{Qualitative result of compositionally.} Compared to the baselines, our method can generate the image with better composition. Specifically, for the last row, Stable Diffusion 3.5 and VSC with 2.1 and 3.5 reflect the prompt better than others, such as silver refrigerator and wood chairs. For the second row, VSC with 2.1 and 3.5 generates the mirror; Stable Diffusion 3.5 fails to reflect the mirror.
    }
    \label{qualitative-exp}
    \vspace{-1em}
\end{figure*}

\Tref{scaling_obj} presents a BLIP-vqa score comparison between two methods, SynGen, which updates the latent with cross-attention loss, and VSC across different numbers of attribute-object binding pairs (3, 4, and 5). For each binding pair configuration, our method outperforms SynGen. 
These results indicate that our method consistently performs better in handling complex attribute-object binding scenarios, with increasing accuracy benefits over SynGen as the complexity of the binding pairs increases.
Additionally, we provide the qualitative results of when the prompts consisting of 3, 4, and 5 different binding pairs, respectively as shown in \Fref{quali-obj-scale}. Throughout the 3 images, the visual appearance of objects is consistent, with only one major change in the layout of the ``red book'' to fit into the image, highlighting the model's capability to generate coherent images while scaling the complexity of attribute-object bindings. 

\subsubsection{Human Eval}
\noindent\textbf{Does VSC generate realistic images?} We supplement this with human evaluations to assess the overall quality of generated images, given that the lack of real reference images makes conventional metrics like the Fréchet Inception Distance (FID) score unsuitable for this task. We assess image quality using Amazon Mechanical Turk (AMT). Raters were given a multiple-choice task, which included four images of the same object cropped from generated images of the same prompt by different models reported in \Tref{human_eval}. We crop objects and compare them individually to discard the correlation of correct binding and visual quality. Raters could also select an option indicating that all images are ``equally good'' or ``equally bad''.  We collected responses from ten raters and recorded the majority decision. If there was no clear majority, the result was categorized as ``no majority winner''. The values represent the fraction of the majority vote from raters, normalized to sum to 100. Furthermore, Figure.\ref{qualitative-exp} shows the qualitative results of compositional generation comparisons.
\begin{table}
    \centering
    \begin{tabular}{c|c}
        \hline
        Methods & Image Quality ($\%$) \\
        \hline 
        Stable Diffusion 3.5 & 12.61 \\
        A\&E & 14.87\\
        SynGen & 25.18\\
        VSC &  \textbf{30.01}\\
        No majority winner & 17.33\\
        \hline 
    \end{tabular}
    \caption{\textbf{Human evaluation on image quality}. Focusing on the visual quality regardless of binding correctness, we crop out the objects and compare them individually. Values are the fraction of the majority vote of raters, normalized to sum to 100. VSC is voted most, implying it generates visually appealing images.} 
    \label{human_eval}
    \vspace{-3em}
\end{table}

Among the methods, VSC received the highest majority vote for visual appeal at 30.01\%, indicating a strong preference among raters. SynGen followed with 25.18\%, while A\&E and Stable Diffusion 3.5 scored lower at 14.87\% and 12.61\%, respectively. Finally, 17.33\% of ratings resulted in no majority winner. The table.~\ref{human_eval} demonstrates that VSC is most favored by raters in terms of realistic quality, outperforming the others, and fine-tuning on synthetic data does not lead to the degeneration of visual quality.
\section{Analysis}

In this section, we examine the impact of scaling the synthetic training dataset, assess the transferability of different compositional types, and evaluate our method's inference cost with an increased number of binding pairs.
\subsection{Scaling Law of Synthetic Data}

\begin{figure}[!htb]
    \centering
    \includegraphics[width=\columnwidth]{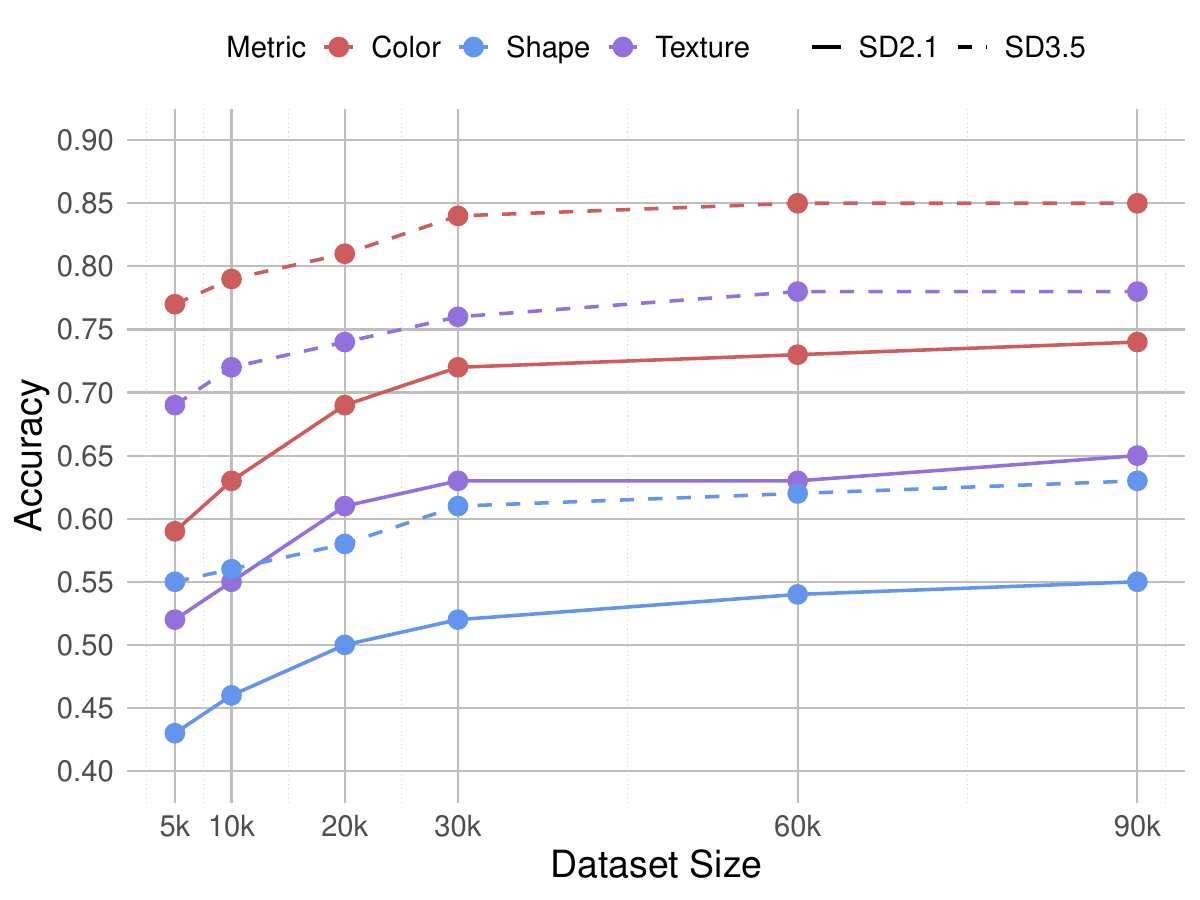}
    \caption{\textbf{Scaling dataset.} The accuracy of models on T2I-CompBench while training on different dataset sizes, highlighting the positive impact of dataset scaling on model performance.}
    \label{scaling}
    \vspace{-1em}
\end{figure}

We trained our model on six different subsets, each containing 5k, 10k, 20k, 30k, 60k, and 90k images, respectively, to analyze the impact of dataset scaling on model performance. As illustrated in \Fref{scaling}, increasing the dataset size leads to a noticeable improvement in accuracy for each attribute, demonstrating that a larger dataset contributes to better model performance. Across all attribute categories—color, texture, and shape—as well as different backbone models, including SD2.1 and SD3.5, the performance steadily improves as the dataset grows from 5,000 to 30,000 images. Beyond this point, further scaling up to 90,000 images provides a marginal enhancement across all categories.



\subsection{Transferability of Compositionality}
To evaluate the model’s ability to generalize compositionality across different groups, we trained it on images from the color group and tested its performance on both the texture and shape groups. 
\begin{table}[htb]
    \centering
    \begin{tabular}{c c c c}
        \hline
        Model & Color & Texture & Shape \\
        \hline
        \multicolumn{4}{c}{Stable Diffusion 2.1}\\
        \hdashline
        Baseline & 0.50 & 0.49 & 0.42 \\
        Full-categories training & 0.74 & \textbf{0.64} & \textbf{0.53} \\
        Color-only training & \textbf{0.77} & 0.60 & 0.46 \\
        \hline
        \multicolumn{4}{c}{Stable Diffusion 3.5} \\
        \hdashline
        Baseline & 0.76 & 0.67 & 0.53 \\
        Full-categories training & 0.85 & \textbf{0.79} & \textbf{0.63} \\
        Color-only training & \textbf{0.89} & 0.74 & 0.58 \\
        \hline
    \end{tabular}
    \caption{\textbf{Transferable compositionality:} We show that training VSC only on ``Color'' category can improve the performance on the remaining categories. Color-only VSC outperforms the baseline on ``Texture'' and ``Shape'', demonstrating the transferability.}
    \label{transferability}
    \vspace{-1em}
\end{table}
\Tref {transferability} compares the performance of two models, a model trained on all compositional categories (``Full-categories training''), a model trained only on the color category (``Color-only training''), and the baseline — across three attributes. In both backbones, the ``Color-only training'' model slightly outperforms the ``Full-categories training'' in Color and importantly shows marginal advances in Texture and Shape compared to the baseline. This outcome suggests that as long as the pre-trained diffusion model can exceptionally generate images from a single pair, VSC can control it in multi-pair settings even if the pair is not in the training set.

\subsection{Inference Time Cost}
\begin{figure}[!htb]
    \centering
    \includegraphics[width=\columnwidth]{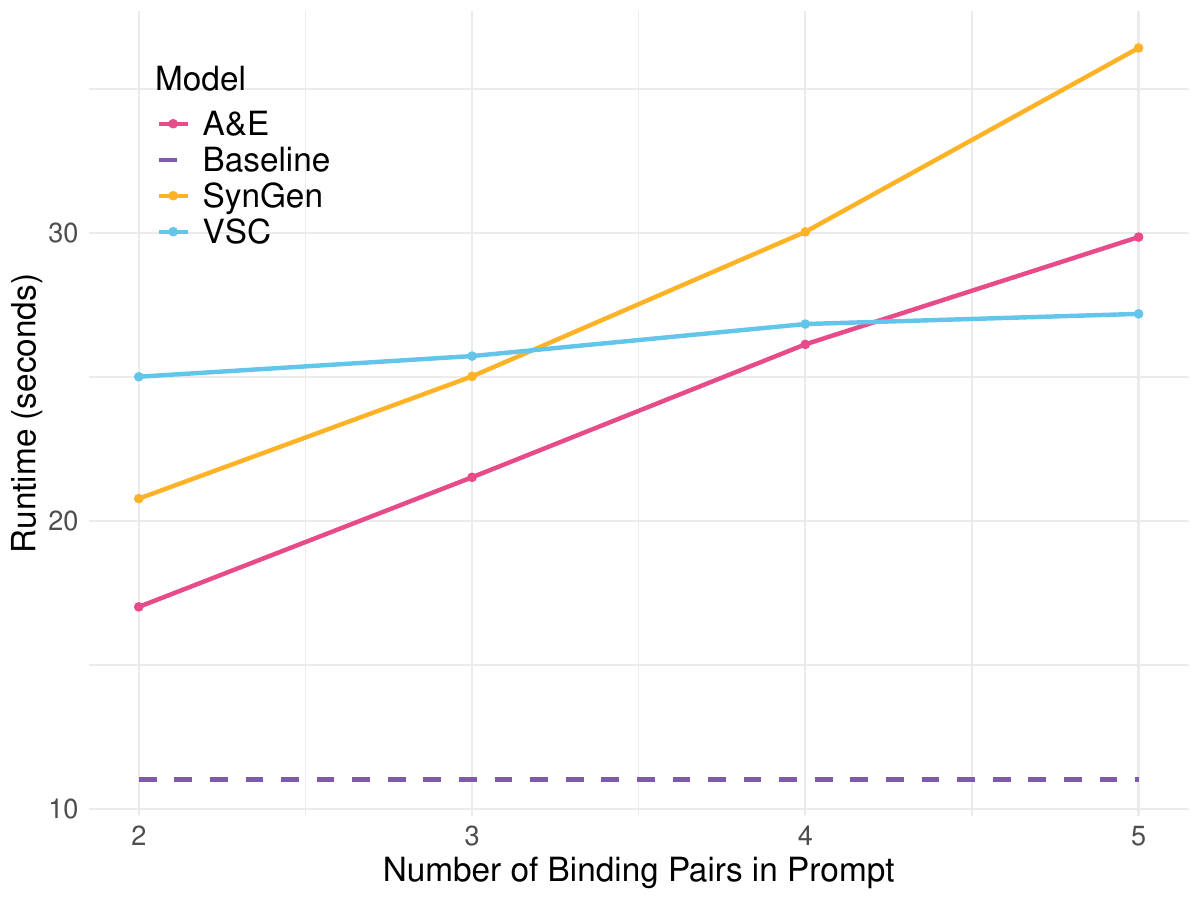}
    \caption{\textbf{Inference cost.} We show the inference time of VSC, SynGen, A\&E, and the baseline according to the number of binding pairs in the prompt. Although VSC is the slowest in 2-pairs generation, VSC efficiently scales to be faster than SynGen and A\&E after the number of pairs exceeds 4 and 5, respectively.}
    \label{inference_time}
    \vspace{-1em}
\end{figure}

VSC efficiently scales its inference time by generating all supporting images in parallel, making it more scalable than SynGen and Attend-and-Excite, and it requires roughly 1.1GB VRAM for each additional reference prompt. As visualized in Figure.\ref{inference_time}, across different numbers of binding pairs, the inference time of the VSC remains approximately 2.26 times (2.26x) the inference time of the default SD 3.5. In contrast, SynGen starts at 2.44x and Attend-and-Excite at 2.28x of the baseline when the prompt contains only two pairs. However, as the number of pairs increases, the inference cost rises significantly for SynGen and Attend-and-Excite, making VSC the fastest model when handling prompts with five pairs, excluding the baseline.
\section{Ablation Studies}

\subsection{Single binding generation}
\label{single-compo}
\begin{table}[!htb]
    \centering
    \begin{tabular}{c|c c c}
        \hline
        Model & Color & Texture & Shape \\
        \hline
        Stable Diffusion 2.1 & 0.9483 & 0.9398 & 0.8520\\
        \hline
    \end{tabular}
    \caption{\textbf{Single attribute-object prompt generation.} We illustrate that the pre-trained Stable Diffusion 2.1 is already superior in generating images from any single pair in the T2I-CompBench.}
    \label{one-binding}
    \vspace{-1em}
\end{table}
We extract all attribute-object pairs in T2I-CompBench \cite{huang2023t2i}, then for each pair, we sample 10 images and finally compute the BLIP-vqa score. 
The results in \Tref{one-binding} demonstrates that SD2.1 already achieves outstanding accuracies on all categories in the single binding setting, discarding the need for data curation to be effectively used as reference images in both training and inference.

\subsection{Image Encoder}

Zarei et al \cite{zarei2024understanding} fine-tuned only an MLP and got a comparable result, so we ablate the importance of fine-tuning some late layers of the image encoder.

\begin{figure}[!htb]
    \centering
    \includegraphics[width=\columnwidth]{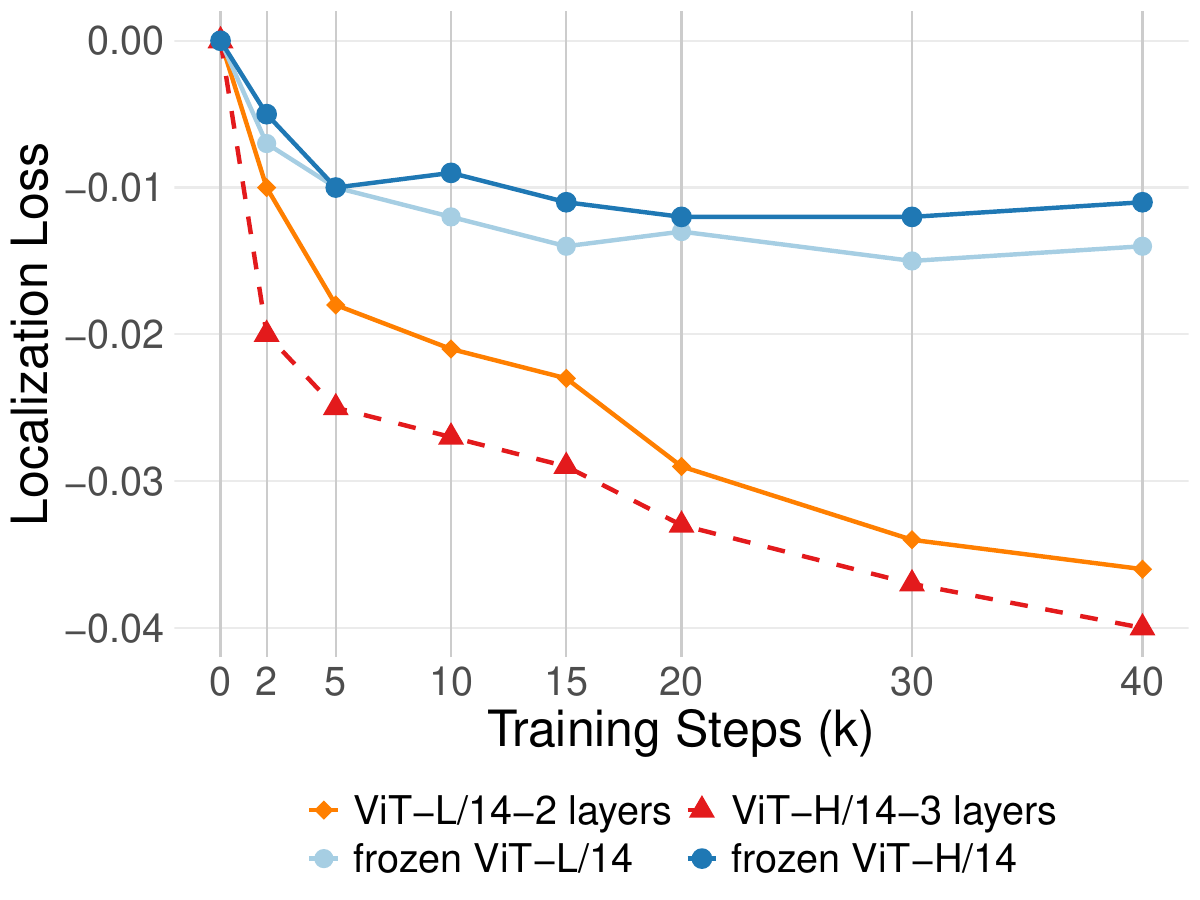}
    \caption{\textbf{Localization cross-attention Loss ($L_{Loc}$).} We depict the importance of fine-tuning the last few layers of the image encoder, as it help the whole pipeline to descent the $L_{Loc}$.}
    \label{abl-img-enc}
    \vspace{-1em}
\end{figure}

\Fref{abl-img-enc} shows the localization loss over training steps for different model configurations. 
The plot compares four setups: ViT-L/14 with trainable 2 layers, ViT-H/14 with 3 trainable layers, and 2 frozen configurations of ViT-L/14 and ViT-H/14. 
Both fine-tuning procedures make the localization loss converge to approximately $-0.04$, while frozen models maintain high loss levels without any significant decrease after the training steps pass $5000$ steps. The need to train the last 2-3 layers of the image encoder may stem from the nature of prototype embeddings. Unlike other subject-driven image generation frameworks, which are optimized to perfectly preserve the unique identity of a specific subject, our focus is on the binding problem and on maintaining diversity in the generated images. Consequently, the cross-attention operation on prototype embeddings is less accurate compared to using exact-object embeddings.

\subsection{Cross-attention Maps}
\label{abl-attn}
To support our previous claim about the high precision of the cross-attention map in Section.~\ref{eval_sec}, we compare the cross-attention maps of the binding tokens obtained from VSC and the pre-trained SD2.1.
\begin{figure}[!htb]
    \centering
    \includegraphics[width=\columnwidth]{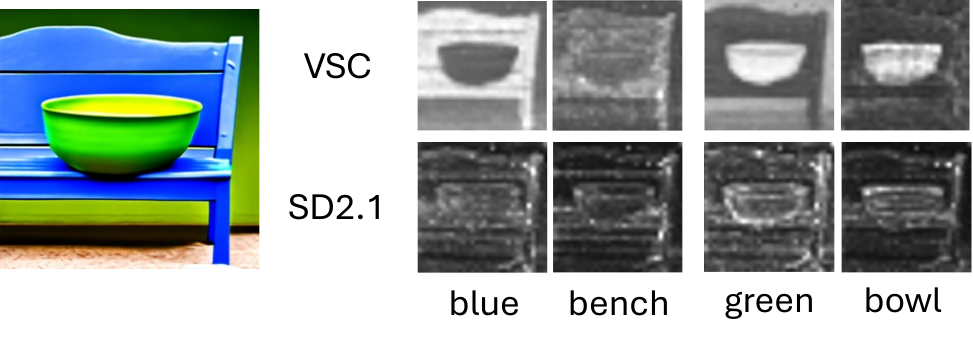}
    \caption{\textbf{Visualization of cross-attention maps.} In VSC, the maps are more fine-grained, and the pairs’ maps are well-aligned.}
    \label{attn}
    \vspace{-1em}
\end{figure}
Given a correct image, we use DDIM to revert the image back to timestep 20, feed-forward to the UNet backbone, and visualize the cross-attention maps of both models at that timestep. Figure.~\ref{attn} showcases cross-attention maps for a given prompt. In the SD2.1, attention maps exhibit inaccuracies, with tokens incorrectly focusing on unrelated pixels. In contrast, in VSC, objects and attributes align more precisely with their corresponding pixels even when the objects are overlapping, assisting our claims.
\section{Conclusion}

This work addresses the critical challenge of enhancing attribute-object binding in recent text-to-image generative models, especially diffusion models, by introducing a novel pairwise visual embedding fusion approach. By decomposing complex text prompts into sub-prompts, generating visual prototypes, and incorporating segmentation-based localization training, our proposed method noticeably improves compositional generalization and accuracy. The results demonstrate superior performance on the T2I CompBench benchmark across various attributes, including color, texture, and shape, even under scaling complexity in binding pairs. Additionally, qualitative and quantitative evaluations validate the method's robustness and adaptability compared to the state-of-the-art models. These findings highlight the potential of combining linguistic and visual embeddings in improving generative models and paving the way for future advancements in creating highly compliant compositional images from complex textual prompts. Future work could explore expanding this approach to broader applications, such as complex scene generation or dynamic object interactions.

{
    \small
    \bibliographystyle{ieeenat_fullname}
    \bibliography{main}

\begin{thebibliography}{36}
\providecommand{\natexlab}[1]{#1}
\providecommand{\url}[1]{\texttt{#1}}
\expandafter\ifx\csname urlstyle\endcsname\relax
  \providecommand{\doi}[1]{doi: #1}\else
  \providecommand{\doi}{doi: \begingroup \urlstyle{rm}\Url}\fi

\bibitem[Agarwal et~al.(2023)Agarwal, Karanam, Joseph, Saxena, Goswami, and Srinivasan]{agarwal2023star}
Aishwarya Agarwal, Srikrishna Karanam, KJ Joseph, Apoorv Saxena, Koustava Goswami, and Balaji~Vasan Srinivasan.
\newblock A-star: Test-time attention segregation and retention for text-to-image synthesis.
\newblock In \emph{Proceedings of the IEEE/CVF International Conference on Computer Vision}, pages 2283--2293, 2023.

\bibitem[Avrahami et~al.(2023)Avrahami, Hayes, Gafni, Gupta, Taigman, Parikh, Lischinski, Fried, and Yin]{avrahami2023spatext}
Omri Avrahami, Thomas Hayes, Oran Gafni, Sonal Gupta, Yaniv Taigman, Devi Parikh, Dani Lischinski, Ohad Fried, and Xi Yin.
\newblock Spatext: Spatio-textual representation for controllable image generation.
\newblock In \emph{Proceedings of the IEEE/CVF Conference on Computer Vision and Pattern Recognition}, pages 18370--18380, 2023.

\bibitem[Campbell et~al.(2024)Campbell, Rane, Giallanza, De~Sabbata, Ghods, Joshi, Ku, Frankland, Griffiths, Cohen, et~al.]{campbell2024understanding}
Declan Campbell, Sunayana Rane, Tyler Giallanza, Nicol{\`o} De~Sabbata, Kia Ghods, Amogh Joshi, Alexander Ku, Steven~M Frankland, Thomas~L Griffiths, Jonathan~D Cohen, et~al.
\newblock Understanding the limits of vision language models through the lens of the binding problem.
\newblock \emph{arXiv preprint arXiv:2411.00238}, 2024.

\bibitem[Chefer et~al.(2023)Chefer, Alaluf, Vinker, Wolf, and Cohen-Or]{chefer2023attend}
Hila Chefer, Yuval Alaluf, Yael Vinker, Lior Wolf, and Daniel Cohen-Or.
\newblock Attend-and-excite: Attention-based semantic guidance for text-to-image diffusion models.
\newblock \emph{ACM Transactions on Graphics (TOG)}, 42\penalty0 (4):\penalty0 1--10, 2023.

\bibitem[Cheng et~al.(2022)Cheng, Misra, Schwing, Kirillov, and Girdhar]{cheng2021mask2former}
Bowen Cheng, Ishan Misra, Alexander~G. Schwing, Alexander Kirillov, and Rohit Girdhar.
\newblock Masked-attention mask transformer for universal image segmentation.
\newblock 2022.

\bibitem[Chomsky(2014)]{chomsky2014aspects}
Noam Chomsky.
\newblock \emph{Aspects of the Theory of Syntax}.
\newblock Number~11. MIT press, 2014.

\bibitem[Dahary et~al.(2024)Dahary, Patashnik, Aberman, and Cohen-Or]{dahary2024yourself}
Omer Dahary, Or Patashnik, Kfir Aberman, and Daniel Cohen-Or.
\newblock Be yourself: Bounded attention for multi-subject text-to-image generation.
\newblock \emph{arXiv preprint arXiv:2403.16990}, 2\penalty0 (5), 2024.

\bibitem[Esser et~al.(2024)Esser, Kulal, Blattmann, Entezari, M{\"u}ller, Saini, Levi, Lorenz, Sauer, Boesel, et~al.]{esser2024scaling}
Patrick Esser, Sumith Kulal, Andreas Blattmann, Rahim Entezari, Jonas M{\"u}ller, Harry Saini, Yam Levi, Dominik Lorenz, Axel Sauer, Frederic Boesel, et~al.
\newblock Scaling rectified flow transformers for high-resolution image synthesis.
\newblock In \emph{Forty-first International Conference on Machine Learning}, 2024.

\bibitem[Feng et~al.(2022)Feng, He, Fu, Jampani, Akula, Narayana, Basu, Wang, and Wang]{feng2022training}
Weixi Feng, Xuehai He, Tsu-Jui Fu, Varun Jampani, Arjun Akula, Pradyumna Narayana, Sugato Basu, Xin~Eric Wang, and William~Yang Wang.
\newblock Training-free structured diffusion guidance for compositional text-to-image synthesis.
\newblock \emph{arXiv preprint arXiv:2212.05032}, 2022.

\bibitem[Feng et~al.(2024)Feng, Zhu, Fu, Jampani, Akula, He, Basu, Wang, and Wang]{feng2024layoutgpt}
Weixi Feng, Wanrong Zhu, Tsu-jui Fu, Varun Jampani, Arjun Akula, Xuehai He, Sugato Basu, Xin~Eric Wang, and William~Yang Wang.
\newblock Layoutgpt: Compositional visual planning and generation with large language models.
\newblock \emph{Advances in Neural Information Processing Systems}, 36, 2024.

\bibitem[Gal et~al.(2022)Gal, Alaluf, Atzmon, Patashnik, Bermano, Chechik, and Cohen-Or]{gal2022image}
Rinon Gal, Yuval Alaluf, Yuval Atzmon, Or Patashnik, Amit~H Bermano, Gal Chechik, and Daniel Cohen-Or.
\newblock An image is worth one word: Personalizing text-to-image generation using textual inversion.
\newblock \emph{arXiv preprint arXiv:2208.01618}, 2022.

\bibitem[Gal et~al.(2023)Gal, Arar, Atzmon, Bermano, Chechik, and Cohen-Or]{gal2023encoder}
Rinon Gal, Moab Arar, Yuval Atzmon, Amit~H Bermano, Gal Chechik, and Daniel Cohen-Or.
\newblock Encoder-based domain tuning for fast personalization of text-to-image models.
\newblock \emph{ACM Transactions on Graphics (TOG)}, 42\penalty0 (4):\penalty0 1--13, 2023.

\bibitem[Huang et~al.(2023)Huang, Sun, Xie, Li, and Liu]{huang2023t2i}
Kaiyi Huang, Kaiyue Sun, Enze Xie, Zhenguo Li, and Xihui Liu.
\newblock T2i-compbench: A comprehensive benchmark for open-world compositional text-to-image generation.
\newblock \emph{Advances in Neural Information Processing Systems}, 36:\penalty0 78723--78747, 2023.

\bibitem[Ilharco et~al.(2021)Ilharco, Wortsman, Wightman, Gordon, Carlini, Taori, Dave, Shankar, Namkoong, Miller, Hajishirzi, Farhadi, and Schmidt]{ilharco_gabriel_2021_5143773}
Gabriel Ilharco, Mitchell Wortsman, Ross Wightman, Cade Gordon, Nicholas Carlini, Rohan Taori, Achal Dave, Vaishaal Shankar, Hongseok Namkoong, John Miller, Hannaneh Hajishirzi, Ali Farhadi, and Ludwig Schmidt.
\newblock Openclip, 2021.
\newblock If you use this software, please cite it as below.

\bibitem[Kirstain et~al.(2023)Kirstain, Levy, and Polyak]{kirstain2023x}
Yuval Kirstain, Omer Levy, and Adam Polyak.
\newblock X\&fuse: Fusing visual information in text-to-image generation.
\newblock \emph{arXiv preprint arXiv:2303.01000}, 2023.

\bibitem[Kumari et~al.(2023)Kumari, Zhang, Zhang, Shechtman, and Zhu]{kumari2023multi}
Nupur Kumari, Bingliang Zhang, Richard Zhang, Eli Shechtman, and Jun-Yan Zhu.
\newblock Multi-concept customization of text-to-image diffusion.
\newblock In \emph{Proceedings of the IEEE/CVF Conference on Computer Vision and Pattern Recognition}, pages 1931--1941, 2023.

\bibitem[Lewis et~al.(2022)Lewis, Nayak, Yu, Yu, Merullo, Bach, and Pavlick]{lewis2022does}
Martha Lewis, Nihal~V Nayak, Peilin Yu, Qinan Yu, Jack Merullo, Stephen~H Bach, and Ellie Pavlick.
\newblock Does clip bind concepts? probing compositionality in large image models.
\newblock \emph{arXiv preprint arXiv:2212.10537}, 2022.

\bibitem[Liu et~al.(2022)Liu, Li, Du, Torralba, and Tenenbaum]{liu2022compositional}
Nan Liu, Shuang Li, Yilun Du, Antonio Torralba, and Joshua~B Tenenbaum.
\newblock Compositional visual generation with composable diffusion models.
\newblock In \emph{European Conference on Computer Vision}, pages 423--439. Springer, 2022.

\bibitem[Meral et~al.(2024)Meral, Simsar, Tombari, and Yanardag]{meral2024conform}
Tuna Han~Salih Meral, Enis Simsar, Federico Tombari, and Pinar Yanardag.
\newblock Conform: Contrast is all you need for high-fidelity text-to-image diffusion models.
\newblock In \emph{Proceedings of the IEEE/CVF Conference on Computer Vision and Pattern Recognition}, pages 9005--9014, 2024.

\bibitem[Mokady et~al.(2023)Mokady, Hertz, Aberman, Pritch, and Cohen-Or]{mokady2023null}
Ron Mokady, Amir Hertz, Kfir Aberman, Yael Pritch, and Daniel Cohen-Or.
\newblock Null-text inversion for editing real images using guided diffusion models.
\newblock In \emph{Proceedings of the IEEE/CVF Conference on Computer Vision and Pattern Recognition}, pages 6038--6047, 2023.

\bibitem[Nie et~al.(2024)Nie, Liu, Mardani, Liu, Eckart, and Vahdat]{nie2024compositional}
Weili Nie, Sifei Liu, Morteza Mardani, Chao Liu, Benjamin Eckart, and Arash Vahdat.
\newblock Compositional text-to-image generation with dense blob representations.
\newblock \emph{arXiv preprint arXiv:2405.08246}, 2024.

\bibitem[Okawa et~al.(2024)Okawa, Lubana, Dick, and Tanaka]{okawa2024compositional}
Maya Okawa, Ekdeep~S Lubana, Robert Dick, and Hidenori Tanaka.
\newblock Compositional abilities emerge multiplicatively: Exploring diffusion models on a synthetic task.
\newblock \emph{Advances in Neural Information Processing Systems}, 36, 2024.

\bibitem[Radford et~al.(2021)Radford, Kim, Hallacy, Ramesh, Goh, Agarwal, Sastry, Askell, Mishkin, Clark, et~al.]{radford2021learning}
Alec Radford, Jong~Wook Kim, Chris Hallacy, Aditya Ramesh, Gabriel Goh, Sandhini Agarwal, Girish Sastry, Amanda Askell, Pamela Mishkin, Jack Clark, et~al.
\newblock Learning transferable visual models from natural language supervision.
\newblock In \emph{International conference on machine learning}, pages 8748--8763. PMLR, 2021.

\bibitem[Rassin et~al.(2024)Rassin, Hirsch, Glickman, Ravfogel, Goldberg, and Chechik]{rassin2024linguistic}
Royi Rassin, Eran Hirsch, Daniel Glickman, Shauli Ravfogel, Yoav Goldberg, and Gal Chechik.
\newblock Linguistic binding in diffusion models: Enhancing attribute correspondence through attention map alignment.
\newblock \emph{Advances in Neural Information Processing Systems}, 36, 2024.

\bibitem[Rombach et~al.(2022)Rombach, Blattmann, Lorenz, Esser, and Ommer]{rombach2022high}
Robin Rombach, Andreas Blattmann, Dominik Lorenz, Patrick Esser, and Bj{\"o}rn Ommer.
\newblock High-resolution image synthesis with latent diffusion models.
\newblock In \emph{Proceedings of the IEEE/CVF conference on computer vision and pattern recognition}, pages 10684--10695, 2022.

\bibitem[Ruiz et~al.(2023)Ruiz, Li, Jampani, Pritch, Rubinstein, and Aberman]{ruiz2023dreambooth}
Nataniel Ruiz, Yuanzhen Li, Varun Jampani, Yael Pritch, Michael Rubinstein, and Kfir Aberman.
\newblock Dreambooth: Fine tuning text-to-image diffusion models for subject-driven generation.
\newblock In \emph{Proceedings of the IEEE/CVF conference on computer vision and pattern recognition}, pages 22500--22510, 2023.

\bibitem[Sarukkai et~al.(2024)Sarukkai, Li, Ma, R{\'e}, and Fatahalian]{sarukkai2024collage}
Vishnu Sarukkai, Linden Li, Arden Ma, Christopher R{\'e}, and Kayvon Fatahalian.
\newblock Collage diffusion.
\newblock In \emph{Proceedings of the IEEE/CVF Winter Conference on Applications of Computer Vision}, pages 4208--4217, 2024.

\bibitem[Shi et~al.(2024)Shi, Xiong, Lin, and Jung]{shi2024instantbooth}
Jing Shi, Wei Xiong, Zhe Lin, and Hyun~Joon Jung.
\newblock Instantbooth: Personalized text-to-image generation without test-time finetuning.
\newblock In \emph{Proceedings of the IEEE/CVF Conference on Computer Vision and Pattern Recognition}, pages 8543--8552, 2024.

\bibitem[Toker et~al.(2024)Toker, Orgad, Ventura, Arad, and Belinkov]{toker2024diffusion}
Michael Toker, Hadas Orgad, Mor Ventura, Dana Arad, and Yonatan Belinkov.
\newblock Diffusion lens: Interpreting text encoders in text-to-image pipelines.
\newblock \emph{arXiv preprint arXiv:2403.05846}, 2024.

\bibitem[Tong et~al.(2024)Tong, Liu, Zhai, Ma, LeCun, and Xie]{tong2024eyes}
Shengbang Tong, Zhuang Liu, Yuexiang Zhai, Yi Ma, Yann LeCun, and Saining Xie.
\newblock Eyes wide shut? exploring the visual shortcomings of multimodal llms.
\newblock In \emph{Proceedings of the IEEE/CVF Conference on Computer Vision and Pattern Recognition}, pages 9568--9578, 2024.

\bibitem[Treisman and Gelade(1980)]{treisman1980feature}
Anne~M Treisman and Garry Gelade.
\newblock A feature-integration theory of attention.
\newblock \emph{Cognitive psychology}, 12\penalty0 (1):\penalty0 97--136, 1980.

\bibitem[Wang et~al.(2024)Wang, Chen, Chen, Ma, Lu, and Lin]{wang2024compositional}
Ruichen Wang, Zekang Chen, Chen Chen, Jian Ma, Haonan Lu, and Xiaodong Lin.
\newblock Compositional text-to-image synthesis with attention map control of diffusion models.
\newblock In \emph{Proceedings of the AAAI Conference on Artificial Intelligence}, pages 5544--5552, 2024.

\bibitem[Wei et~al.(2023)Wei, Zhang, Ji, Bai, Zhang, and Zuo]{wei2023elite}
Yuxiang Wei, Yabo Zhang, Zhilong Ji, Jinfeng Bai, Lei Zhang, and Wangmeng Zuo.
\newblock Elite: Encoding visual concepts into textual embeddings for customized text-to-image generation.
\newblock In \emph{Proceedings of the IEEE/CVF International Conference on Computer Vision}, pages 15943--15953, 2023.

\bibitem[Xiao et~al.(2024)Xiao, Yin, Freeman, Durand, and Han]{xiao2024fastcomposer}
Guangxuan Xiao, Tianwei Yin, William~T Freeman, Fr{\'e}do Durand, and Song Han.
\newblock Fastcomposer: Tuning-free multi-subject image generation with localized attention.
\newblock \emph{International Journal of Computer Vision}, pages 1--20, 2024.

\bibitem[Zarei et~al.(2024)Zarei, Rezaei, Basu, Saberi, Moayeri, Kattakinda, and Feizi]{zarei2024understanding}
Arman Zarei, Keivan Rezaei, Samyadeep Basu, Mehrdad Saberi, Mazda Moayeri, Priyatham Kattakinda, and Soheil Feizi.
\newblock Understanding and mitigating compositional issues in text-to-image generative models.
\newblock \emph{arXiv preprint arXiv:2406.07844}, 2024.

\bibitem[Zheng et~al.(2023)Zheng, Zhou, Li, Qi, Shan, and Li]{zheng2023layoutdiffusion}
Guangcong Zheng, Xianpan Zhou, Xuewei Li, Zhongang Qi, Ying Shan, and Xi Li.
\newblock Layoutdiffusion: Controllable diffusion model for layout-to-image generation.
\newblock In \emph{Proceedings of the IEEE/CVF Conference on Computer Vision and Pattern Recognition}, pages 22490--22499, 2023.

\end{thebibliography}
}

\end{document}